\documentclass[11pt,a4paper]{article}
\usepackage[hyperref]{emnlp2020}
\usepackage{times}
\usepackage{latexsym}
\usepackage{graphicx}

\usepackage{amsmath}
\usepackage{amsfonts}
\usepackage{amssymb}
\usepackage{multirow}
\usepackage{tabularx}
\usepackage{booktabs,dcolumn,caption}
\usepackage{comment}
\usepackage{lscape}
\usepackage{xspace}

\definecolor{lightblue}{RGB}{135,206,250}

\usepackage{microtype}

\aclfinalcopy 
\def\aclpaperid{3517} 

\title{Efficient One-Pass End-to-End Entity Linking for Questions}

\author{
    Belinda Z. Li$^\spadesuit$\thanks{\hspace{.06in}Work done while at Facebook AI.}\qquad Sewon Min$^\diamondsuit$ \\
    \textbf{Srinivasan Iyer$^\dagger$\quad Yashar Mehdad$^\dagger$\quad Wen-tau Yih$^\dagger$} \\
    $^\spadesuit$MIT CSAIL \qquad
    $^\diamondsuit$University of Washington \qquad $^\dagger$Facebook AI  \\
    \texttt{bzl@mit.edu} \qquad \texttt{sewon@cs.washington.edu} \\ \texttt{\{sviyer,mehdad,scottyih\}@fb.com}
 }

\newcommand\ignore[1]{}

\newcommand{\red}[1]{\colorbox{pink}{\textbf{#1}}}
\newcommand{\blue}[1]{\colorbox{lime}{\textbf{#1}}}

\newcommand\ourwebq{WebQSP$_\text{EL}$\xspace}
\newcommand\ourgraphq{GraphQ$_\text{EL}$\xspace}
\newcommand\tagme{TAGME}

\DeclareRobustCommand 
      \Compactcdots{\mathinner{\cdotp\mkern-2mu\cdotp\mkern-2mu\cdotp}}

\date{}

\begin{document}
\maketitle
\begin{abstract}

We present ELQ, a fast end-to-end entity linking model for questions, which uses a biencoder to jointly perform mention detection and linking in one pass.
Evaluated on WebQSP and GraphQuestions with extended annotations that cover multiple entities per question, ELQ outperforms the previous state of the art by a large margin of +12.7\% and +19.6\% F1, respectively. With a very fast inference time (1.57 examples/s on a single CPU), ELQ can be useful for downstream question answering systems. In a proof-of-concept experiment, we demonstrate that using ELQ significantly improves the downstream QA performance of GraphRetriever~\cite{min2019knowledge}.\footnote{Code and data available at \url{https://github.com/facebookresearch/BLINK/tree/master/elq}}

\end{abstract}

\section{Introduction}
\label{sec:intro}

Entity linking (EL), the task of identifying entities and mapping them to the correct entries in a database, is crucial for analyzing factoid questions and for building robust question answering (QA) systems.
For instance, the question ``\emph{when did shaq come to the nba?}" can be answered by examining \emph{Shaquille O'Neal}'s Wikipedia article~\cite{min2019knowledge}, or its properties in a knowledge graph~\cite{yih-etal-2015-semantic,yu-etal-2017-improved}. 
However, real-world user questions are invariably noisy and ill-formed, lacking cues provided by casing and punctuation, 
which prove challenging to current end-to-end entity linking systems~\cite{yang-chang-2015-mart,sorokin-gurevych-2018-mixing}. 
While recent pre-trained models have proven highly effective for entity linking~\cite{logeswaran-etal-2019-zero,wu2019zeroshot}, they are only designed for entity disambiguation and require mention boundaries to be given in the input. 
Additionally, such systems have only been evaluated on long, well-formed documents like news articles~\cite{ji2010overview-kbptac},
but not on short, noisy text.
Also, most prior works have focused mainly on improving model prediction accuracy, largely overlooking efficiency.

\begin{figure}[t]
    \centering
    \resizebox{\columnwidth}{!}{
    \includegraphics[width=\textwidth]{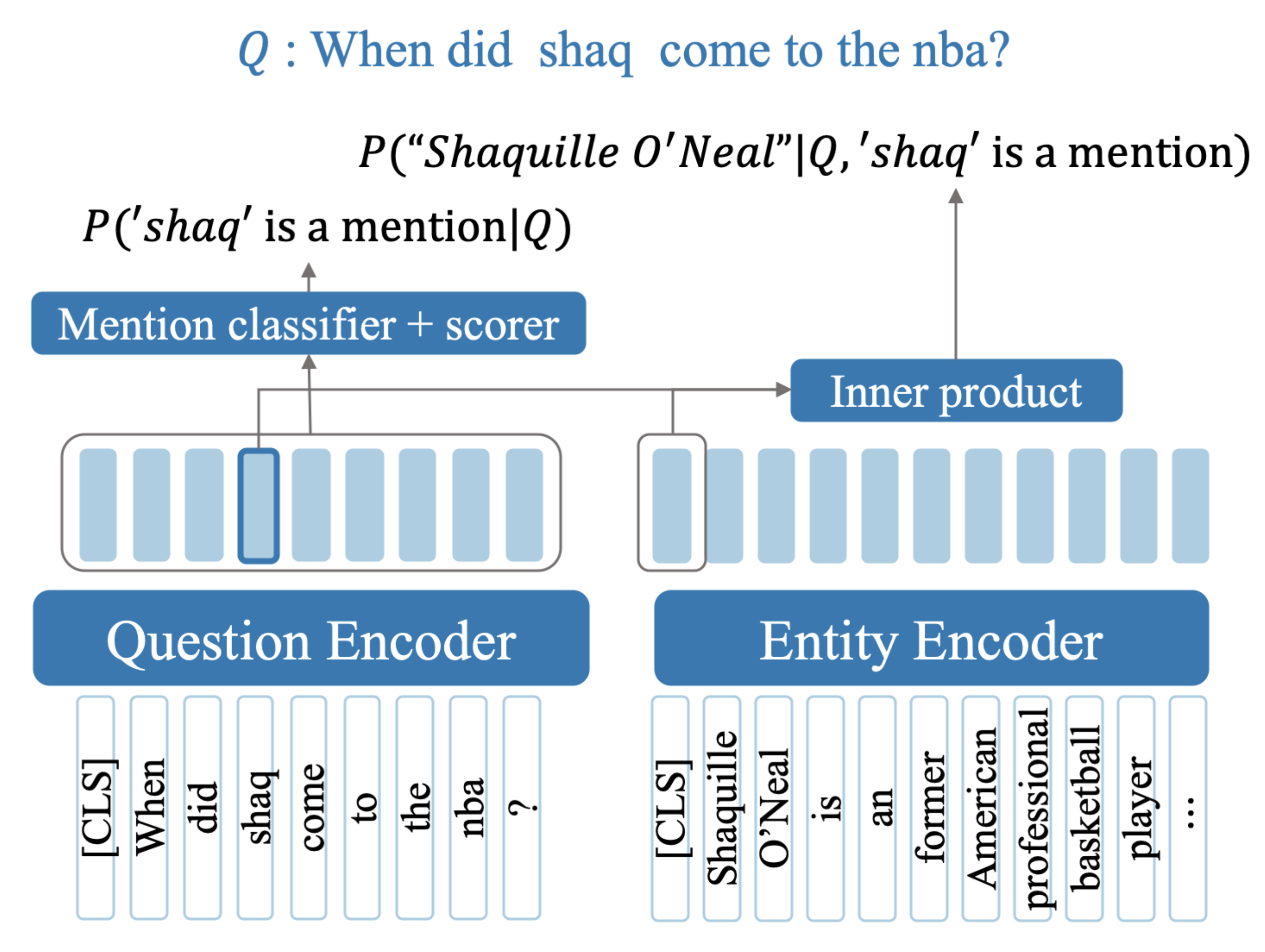}}
    \caption{Overview of our end-to-end entity linking system. We separately encode the question and entity. We use the question representations to jointly detect mentions and score candidate entities through inner-product with the entity vector.
    }
    \label{fig:my_model_2}
\end{figure}

In this work, we propose ELQ, a fast and accurate entity linking system that specifically targets questions.
Following the Wikification setup~\cite{ratinov-etal-2011-local}, ELQ aims to identify the mention boundaries of entities in a given question and their corresponding Wikipedia entity.
We employ a biencoder based on BERT~\cite{devlin2019bert} as shown in Figure~\ref{fig:my_model_2}.
The entity encoder computes entity embeddings for all entities in Wikipedia, using their short descriptions.
Then, the question encoder derives token-level embeddings for the input question.
We detect mention boundaries using these embeddings, and disambiguate each entity mention based on an inner product between the mention embeddings (averaged embedding over mention tokens) and the entity embeddings.
Our model extends the work of \citet{wu2019zeroshot} but with one major difference: our system does not require pre-specified mention boundaries in the input, and is able to jointly perform mention detection and entity disambiguation in just one pass of BERT.
Thus, at inference time, we are able to identify multiple entities in the input question efficiently.

We extend entity disambiguation annotations from \citet{sorokin-gurevych-2018-mixing} to create an end-to-end question entity linking benchmark.
Evaluated on this benchmark, we are able to outperform previous methods in both accuracy and run-time.
ELQ has much faster end-to-end inference time than any other neural baseline (by $2\times$), while being more accurate than all previous models we evaluate against, suggesting that it is practically useful for downstream QA systems.
We verify the applicability of ELQ to practical QA models in a proof-of-concept experiment, by augmenting GraphRetriever~\citep{min2019knowledge} to use our model, improving its downstream QA performance on three open-domain QA datasets (by up to $6\%$).

\section{Related Work}
Much prior work on entity linking has focused on long, grammatically coherent documents that contain many entities.
This setting does not accurately reflect the difficulties of entity linking on questions.
While there has been some previous work on entity linking for questions~\cite{sorokin-gurevych-2018-mixing,10.1145/2684822.2685317,10.1145/3269206.3271809,tan-etal-2017-entity}, such works (mostly from the pre-BERT era) utilize complex models with many interworking modules.
For example,~\newcite{sorokin-gurevych-2018-mixing} proposes a variable-context granularity (VCG) model to address the 
noise and lack of context in questions, which incorporates signals from various levels of granularity by using character-level, token-level, and knowledge-base-level modules.
They also rely on external systems 
as a part of the modeling pipeline.

In this work, we take a much simpler approach that uses a biencoder. Biencoder models have been used in a wide range of tasks~\citep{seo2019real,karpukhin2020dense,wu2019zeroshot}. 
They enable fast inference time through maximum inner product search.
Moreover, as we find, biencoders can be decomposed into reusable question and entity encoders, and we can greatly expedite training by training one component independently of the other.


\section{Problem Definition \& ELQ Model}
\label{sec:model}
We formally define our entity linking task as follows. Given a question $q$ and a set of entities $\mathcal{E} = \{e_i\}$ from Wikipedia, each with titles $t(e_i)$ and text descriptions $d(e_i)$, our goal is to output a list of tuples, $(e, [m_\mathrm{s}, m_\mathrm{e}])$, whereby $e \in \mathcal{E}$ is the entity corresponding to the mention span from the $m_\mathrm{s}$-th to $m_\mathrm{e}$-th token in $q$. In practice, we take the title and first 128 tokens of the entity's Wikipedia article as its title $t(e_i)$ and description $d(e_i)$.

We propose an end-to-end entity linking system that performs both mention detection and entity disambiguation on questions in one pass of BERT.

Given an input question $q = q_1 \Compactcdots q_n$ of length $n$, we first obtain question token representations based on BERT~\citep{devlin2019bert}:
\begin{equation*}
    [\mathbf{q}_1 \Compactcdots \mathbf{q}_n]^{\intercal} = \mathrm{BERT}(\mathtt{[CLS]}~q_{1} \Compactcdots q_{n}\mathtt{[SEP]}) \in \mathbb{R}^{n \times h},
\end{equation*}
where each $\mathbf{q_i}$ is a $h$-dimensional vector.
We then obtain entity representations $\mathbf{x}_e$ for every $e_i \in \mathcal{E}$.
\begin{eqnarray*}
    \mathbf{x}_e = \mathrm{BERT}_{_\mathtt{[CLS]}}(\mathtt{[CLS]}t(e_i) \mathtt{[ENT]}d(e_i)\mathtt{[SEP]}) \in \mathbb{R}^h,
\end{eqnarray*}
where ${\mathtt{[CLS]}}$ indicates that we select the representation of the $\mathtt{[CLS]}$ token.
We consider candidate mentions as all spans $[i, j]$ ($i$-th to $j$-th tokens of $q$) in the text up to length $L$.

\paragraph{Mention Detection}
To compute the likelihood score of a candidate span $[i, j]$ being an entity mention, we first obtain scores for each token being the start or the end of a mention:
$$
    s_\text{start}(i) = \mathbf{w}_\mathrm{start}^{\intercal} \mathbf{q}_i, \quad
    s_\text{end}(j) = \mathbf{w}_\mathrm{end}^{\intercal} \mathbf{q}_j,
$$ where $\mathbf{w}_\mathrm{start}, \mathbf{w}_\mathrm{end} \in \mathbb{R}^h$ are learnable vectors.
We additionally compute scores for each token $t$ being part of a mention:
\begin{eqnarray*}
    s_\text{mention}(t) &=& \mathbf{w}_\mathrm{mention}^{\intercal} \mathbf{q}_t,
\end{eqnarray*} where $\mathbf{w}_\mathrm{mention} \in \mathbb{R}^h$ is a learnable vector.
We finally compute mention probabilities as: 
\begin{equation*}
    p([i,j]) = \sigma(s_\text{start}(i) + s_\text{end}(j) + \sum_{t=i}^j s_\text{mention}(t)).
\end{equation*}

\paragraph{Entity Disambiguation} 
We obtain a mention representation for each mention candidate $[i,j]$ by averaging $\mathbf{q_i} \Compactcdots \mathbf{q_j}$, and compute a similarity score $s$ between the mention candidate and an entity candidate $e \in \mathcal{E}$:
\begin{eqnarray*}
    \textbf{y}_{i,j} &=& \frac{1}{(j - i + 1)}\sum_{t=i}^j \mathbf{q}_t \in \mathbb{R}^h, \\
    s(e, [i,j]) &=& \mathbf{x}_e^{\intercal} \mathbf{y}_{i,j}.
\end{eqnarray*}

We then compute a likelihood distribution over all entities, conditioned on the mention $[i, j]$:
\begin{eqnarray*}
    \label{eq:entity_p}
    p(e|[i,j]) = \frac{\text{exp}(s(e, [i,j]))}{\sum_{e'\in \mathcal{E}} \text{exp}(s(e', [i,j]))}.
\end{eqnarray*}

\paragraph{Training}
We jointly train the mention detection and entity disambiguation components by optimizing the sum of their losses. 
We use a binary cross-entropy loss across all mention candidates:
\begin{eqnarray*}
    \mathcal{L}_\text{MD} &=& -\frac{1}{N} \sum_{\substack{1 \leq i \leq j \leq \\ \min(i + L - 1, n)}}\Big(y_{[i,j]} \log p([i,j]) \\
    && + (1 - y_{[i,j]}) \log{(1 - p([i,j]))}\Big) ,
\end{eqnarray*}
whereby $y_{[i,j]} = 1$ if $[i,j]$ is a gold mention span, and $0$ otherwise. $N$ is the total number of candidates we consider.\footnote{If $n\ge L$, $N = L(L+1) / 2 + (n - L)L$. Otherwise, $N = n(n+1) / 2$.}

The entity disambiguation loss is given by
\begin{equation*}
    \mathcal{L}_\text{ED} = - \log{p(e_g | [i,j])},
\end{equation*}
where $e_g$ is the gold entity corresponding to mention $[i,j]$.

To expedite training, we use a simple transfer learning technique:
we take the entity encoder trained on Wikipedia by \citet{wu2019zeroshot} and freeze its weights, training only the question encoder on QA data.
In addition, we mine hard negatives. 
As entity encodings are fixed, a fast search of hard negatives in real time is possible.

\paragraph{Inference}
Figure~\ref{fig:my_model_2} shows our inference process.
Given an input question $q$, we use our mention detection model to obtain our mention set $\mathcal{M} = \{[i, j]: 1 \leq i \leq j \leq \min(i+L-1, n), p([i,j]) > \gamma \}$, where $\gamma$ is our threshold (a hyperparameter).
We then compute $p(e, [i,j]) = p(e | [i,j])p([i,j])$ for each mention $[i, j] \in \mathcal{M}$, and threshold according to $\gamma$.
In contrast to a two-stage pipeline which first extracts mentions, then disambiguates entities~\cite{fevry2020empirical}, a joint approach grants us the flexibility to consider multiple possible candidate mentions for entity linking. This can be crucial in questions as it can be difficult to 
extract mentions from short, noisy text in a single step.

More implementation details can be found in Appendix~\ref{app:impl-details}.

\section{Experiments}
\label{sec:experiments}
\begin{table}[t]
    \small
    \centering
    \begin{tabular}{rcccc}
    \toprule
        \multirow{2}{*}{\textbf{Data}} & \multicolumn{2}{c}{\textbf{Train}} &  \multicolumn{2}{c}{\textbf{Test}}  \\
        \cmidrule(lr){2-3} \cmidrule(lr){4-5}
         & \textbf{\#Q} & \textbf{\#E}  & \textbf{\#Q} & \textbf{\#E} \\
    \midrule
        \ourwebq\ & 2974 & 3242 & 1603 & 1806 \\
        \ourgraphq\ & 2089 & 2253 & 2075 & 2229 \\
    \bottomrule
    \end{tabular}
    \caption{Dataset statistics of \ourwebq\ and \ourgraphq. \#Q and \#E indicate the number of questions and entities, respectively.}
    \label{tab:data}
\end{table}

\begin{table*}
    \small
    \centering
    \begin{tabular}{rrcccccccc}
    \toprule
        \multirow{2}{*}{\textbf{Training Data}} &
        \multirow{2}{*}{\textbf{Model}} & \multicolumn{4}{c}{\textbf{\ourwebq}} &  \multicolumn{4}{c}{\textbf{\ourgraphq (zero-shot)}}  \\
        \cmidrule(lr){3-6} \cmidrule(lr){7-10}
        & & \textbf{Prec} & \textbf{Recall} & \textbf{F1} & \textbf{\#Q/s} & \textbf{Prec} & \textbf{Recall} & \textbf{F1} & \textbf{\#Q/s} \\
    \midrule
        \multirow{2}{*}{\ourwebq} & VCG$^\dagger$ & 82.4 & 68.3 & 74.7 & 0.45 & 54.1 & 30.6 & 39.0 & 0.26 \\
        & ELQ & \underline{90.0} & \underline{85.0} & \underline{87.4} & \underline{1.56} & \underline{60.1} & \underline{57.2} & \underline{58.6} & \underline{1.57} \\
    \midrule
        \multirow{3}{*}{Wikipedia} & \tagme & 53.1 & 27.3 & 36.1 & \underline{\textbf{2.39}} & 49.6 & 36.5 & 42.0 & \underline{\textbf{3.16}} \\
        & BLINK & 82.2 & 79.4 & 80.8 & 0.80 & 65.3 & 61.2 & 63.2 & 0.78  \\
        & ELQ & \underline{86.1} & \underline{81.8} & \underline{83.9} & 1.56 & \underline{69.8} & \underline{\textbf{69.8}} & \underline{69.8} & 1.57 \\
    \midrule
        Wikipedia + \ourwebq & ELQ & \textbf{91.0} & \textbf{87.0} & \textbf{89.0} & 1.56 & \textbf{74.7} & 66.4 & \textbf{70.3} & 1.57 \\
    \bottomrule
    \end{tabular}
    \caption{
        Results on \ourwebq\ and \ourgraphq\ test data, under 3 training settings.
        `\#Q/s' (number of questions per second) indicates inference speed on 1 CPU. 
        Models trained in comparable settings are clustered together. Overall highest scores are \textbf{bolded}, while highest scores per setting are \underline{underlined}. 
        \\
        \small $^\dagger$VCG results are different from numbers in the original paper as the evaluation sets are slightly different.
    }
    \label{tab:main_result}
\end{table*}

\subsection{Data}
\label{subsec:dataset}
We evaluate our approach on two QA datasets, WebQSP~\citep{yih2016value} and GraphQuestions~\citep{su2016generating}, with additional entity annotations provided by~\citet{sorokin-gurevych-2018-mixing}.
The original datasets do not have all mention boundary labels annotated. Therefore, in order to evaluate both mention detection and entity disambiguation, we extend previous labels and create new end-to-end question entity-linking datasets, \ourwebq\ and 
\ourgraphq.\footnote{Data available at \url{http://dl.fbaipublicfiles.com/elq/EL4QA_data.tar.gz}}
In line with our task definition, all entities presented in each question are labeled with $(e, [m_\mathrm{s}, m_\mathrm{e}])$, whereby $e \in \mathcal{E}$ is the entity corresponding to the mention span from the $m_\mathrm{s}$-th to $m_\mathrm{e}$-th token in $q$.
We ask four in-house annotators to identify corresponding mention boundaries, given gold entities in the questions.
We exclude examples that link to \verb|null| or no entities,
that are not in Wikipedia, or are incorrect or overly generic (e.g. linking a concept like {\em marry}). 
To check inter-annotation agreement amongst the 4 annotators, we set aside a shared set of documents (comprised of documents from both datasets) that all 4 annotators annotated. We found exact-match inter-annotator agreement to be 95\% (39/41) on this shared set. 

Table~\ref{tab:data} reports the statistics of the resulting datasets, \ourwebq\ and \ourgraphq. Following \newcite{sorokin-gurevych-2018-mixing}, we use \ourwebq\ for training and \ourgraphq\ for zero-shot evaluation.

\paragraph{Evaluation Metrics}
Using the rule defined by \citet{10.1145/2701583.2701591}, a prediction is correct only if the groundtruth entity is identified \textit{and} the predicted mention boundaries overlap with the groundtruth boundaries. (This is sometimes known as ``weak matching".)
Specifically, let ${\mathcal{T}}$ be a set of gold entity-mention tuples and $\widehat{\mathcal{T}}$ be a set of predicted entity-mention tuples, we define precision ($p$), recall ($r$) and F1-score ($F_1$) as follows:
\begin{eqnarray*}\small
    \mathcal{C} &=& \big\{
        e \in \mathcal{E}| [m_s, m_e] \cap [\widehat{m}_s, \widehat{m}_e] \neq \varnothing,
         \\
        && (e, [m_s, m_e]) \in \mathcal{T}, (e, [\widehat{m}_s, \widehat{m}_e]) \in \widehat{\mathcal{T}}
    \big\}, \\
\end{eqnarray*}\vspace{-2.5em}
$$
    p = \frac{|\mathcal{C}|}{|\widehat{\mathcal{T}}|}, \quad
    r = \frac{|\mathcal{C}|}{|\mathcal{T}|}, \quad
    F_1 = \frac{2pr}{p + r}.
$$

\paragraph{Baselines}
We use the following baselines:
(1) \textbf{\tagme}~\cite{6035657}, a lightweight, on-the-fly entity linking system that is popular for many downstream QA tasks, being much faster than most neural models~\citep{joshi2017triviaqa,sun2018open,min2019knowledge}, 
(2) \textbf{VCG}~\citep{sorokin-gurevych-2018-mixing}, the current  state-of-the-art entity linking system on WebQSP, and (3) biencoder from \textbf{BLINK}~\citep{wu2019zeroshot}. 
As BLINK requires pre-specified mention boundaries as input, we train a separate, BERT-based span extraction model on WebQSP in order to predict mention boundaries (details in Appendix~\ref{sec:span_extraction}).

\subsection{Results}\label{subsec:results}
Table~\ref{tab:main_result} show our main results.
We find that BERT-based biencoder models far outperform the state-of-the-art (VCG) on both datasets, 
in performance and in runtime. Moreover, ELQ outperforms all other models trained in a comparable setting, and is much more efficient than every other neural baseline (VCG and BLINK).
ELQ is also up to $2.3\times$ better than TAGME --- in the case of \ourwebq.

\paragraph{Performance} ELQ outperforms BLINK, suggesting that it is possible to train representations from a single model 
to resolve both entity references as well as mention boundaries of all entities in text, without restricting the model to focusing on a single marked entity as in BLINK.

\paragraph{Runtime} We record the inference speed on CPUs in number of questions processed per second for all models (Table~\ref{tab:main_result}). 
For BLINK, we report the combined speed of our span extraction model and the BLINK entity linker, in order to compare the \textit{end-to-end} speeds. 
ELQ, which performs both detection and disambiguation in one pass of BERT, is approximately $2\times$ 
faster than BLINK, which performs multiple passes, while also outperforming BLINK in F1 score.
Moreover, against \tagme~\cite{6035657}, ELQ is only $1.5\times$ slower on \ourwebq and $2.0\times$ slower on \ourgraphq, despite TAGME being a completely non-neural model (with much lower accuracy).

\section{QA Experiments}
\label{sec:qa_experiments}
\begin{table}[t]
    \small
    \centering
    \begin{tabular}{rccc}
    \toprule
        & \textbf{WQ} & \textbf{NQ} & \textbf{TQA} \\
    \midrule
        TF-IDF$^\dagger$ & 20.8 & 28.7 & 54.0 \\
        TAGME + GRetriever$^\dagger$ & 31.8 & 33.5 & 55.0 \\
        ELQ$_\text{Wiki}$ + GRetriever & 37.4 & \textbf{37.4} & \textbf{55.4} \\
        ELQ$_\text{QA}$ + GRetriever & \textbf{37.7} & 37.0 & 54.7 \\
    \bottomrule
    \end{tabular}
    \caption{
        QA result (Exact Match) on the test set of WebQuestions (WQ), Natural Questions (NQ) and TriviaQA (TQA).
        ELQ$_\text{Wiki}$ represents our model trained on Wikipedia data, while ELQ$_\text{QA}$ represents our model trained on Wikipedia+\ourwebq data. \\
        \small $^\dagger$Result taken from \citep{min2019knowledge}.
    }
    \label{tab:qa_results}
\end{table}


To demonstrate the impact of improved entity linking on the end QA accuracy, we experiment with the task of textual open-domain question answering, using GraphRetriever (GRetriever)~\citep{min2019knowledge}.
GRetriever uses entity linking to construct a graph of passages in the retrieval step and deploys a reader model to answer the question.
The original model uses TAGME for entity linking; we replace TAGME with ELQ and keep the other components the same, in order to 
isolate the impact of entity linking.\footnote{\citet{min2019knowledge} used two reader models, ParReader++ and GraphReader; for simplicity, we only use ParReader++}
As an additional baseline, we also add the result of TF-IDF, implemented by~\citet{chen2017reading}, a widely used retrieval system.

Results are shown in Table~\ref{tab:qa_results}.
Following literature in open-domain QA, we evaluate our approach on three datasets, WebQuestions~\citep{berant2013semantic}, Natural Questions~\citep{kwiatkowski2019natural} and TriviaQA~\citep{joshi2017triviaqa}.
In particular, WebQuestions (WQ) and Natural Questions (NQ) consist of short, noisy questions from Web queries, in line with the motivation of our work.
We observe that simply replacing TAGME with ELQ significantly improves performance, including 5.9\% and 3.9\% absolute improvements on WQ and NQ, respectively. 
While ELQ trained on Wikipedia achieves good results overall, further fine-tuning on~\ourwebq gives extra gains on WQ. This indicates that, if entity linking annotations in the same domain are available, using them to finetune ELQ can bring further gains.

\section{Analysis}
\label{sec:analysis}
\begin{table}[t]
    \centering
    \small
    \begin{tabular}{rcc}
    \toprule
        \textbf{Experiment} &
        \textbf{ELQ (Wiki)} & \textbf{BLINK} \\
    \midrule
        MD + EL & 87.3 & 82.2 \\
        MD only & 94.6 & 92.9 \\
        EL only & 90.2 & 86.6 \\
    \bottomrule
    \end{tabular}
    \caption{Analyzing the performance of the mention detector and entity linker respectively on \ourwebq (dev). We compare our Wikipedia-trained model to BLINK. MD + EL refers to the end-to-end F1 score (the normal setup).}
    \label{tab:perf_analysis}
\end{table}

\paragraph{Mention Detector vs. Entity Linker} We set up experiments to disentangle the capability of ELQ's entity linker and mention detector. First, to test just the mention detector (MD only), we measure just the mention boundary overlap between predicted and groundtruth mentions, ignoring the entity label. Next, to test just the entity linker (EL only), we give the entity linking component \textit{gold} mention boundaries, and compute the resulting F1 score. 
We do this for both ELQ and BLINK. For comparability, we use the version of ELQ trained on Wikipedia. Results are reported in Table~\ref{tab:perf_analysis}. Surprisingly, we find that \textit{both} components of ELQ outperform BLINK, 
suggesting that the two tasks might \textit{mutually benefit} from being trained jointly.

\paragraph{Runtime} To confirm that our biencoder's main bottleneck is the BERT forward pass --- and thus, investing in decreasing the number of BERT forward passes is valuable --- we separately time each component of ELQ during inference.
We run examples from \ourwebq test set one at a time through ELQ, on 1 CPU, and average runtimes across all examples. Indeed, we find that the BERT forward pass to be the slowest component of the model, taking $0.683$s, over $6\times$ slower than the next slowest component of the model, inner-product search (taking $0.107$s). Everything else takes a combined total of $5.08\times 10^{-3}$s.


\paragraph{Qualitative} We manually examine all our model's errors on the \ourwebq and \ourgraphq dev sets.
We identify four broad error categories: (1) technically correct --- where our model was technically correct but limitations in evaluation falsely penalized our model (i.e., we found a more or less precise version of the same entity), (2) not enough entities --- where the model did not fully identify all entities in the question, (3) wrong entities --- where our model linked to the wrong entity, (4) insufficient context --- where the model made reasonable mistakes due to the lack of context (that even reasonable humans would make). Error type breakdowns can be found in Table~\ref{tab:qual_analysis}.

\begin{table}[t]
    \centering
    \small
    \begin{tabular}{rcc}
    \toprule
        \textbf{Error Type} & \textbf{\ourwebq} & \textbf{\ourgraphq} \\
    \midrule
        Technically Correct & $49.2$ & $23.3$ \\
        Not Enough Entities & $13.1$ & $51.8$ \\
        Wrong Entities & $26.2$ & $20$ \\
        Insufficient Context & $11.5$ & $5$ \\
    \bottomrule
    \end{tabular}
    \caption{Breakdown of frequency of each error type on each dev set (in terms of \% of all errors on that dataset). We use ELQ trained on Wikipedia +~\ourwebq~here.}
    \label{tab:qual_analysis}
\end{table}

\section{Conclusion}
\label{sec:conclusion}
We proposed an end-to-end model for entity linking on questions that jointly performs mention detection and disambiguation with one pass through BERT.
We showed that it is highly efficient, 
and that it outperforms previous state-of-the-art models on two benchmarks.
Furthermore, when applied to a QA model, ELQ improves that model's end QA accuracy.
Despite being originally designed with questions in mind, we believe ELQ could also generalize to longer, well-formed documents.

\bibliography{emnlp2020}
\bibliographystyle{acl_natbib}

\clearpage
\appendix
\section{Annotation Statistics}
We show annotators a question and a gold entity in the question, and instruct them to annotate (i.e., put brackets around) the appropriate mention span. 

For quality control, we created a shared annotation set by pulling a subset of examples from each dataset, and having all four annotators label that set.
Inter-annotator agreement statistics on our shared set are shown in Table~\ref{tab:data_agreements}. Note that only 2 out of 41 shared examples did not have unanimous mention boundary agreement. The two conflicting examples, with the respective annotations, are shown below:

\begin{center}
\small
\begin{tabular}{l}
\toprule
Question: `who are the two state senators of georgia?' \\
Entity: `United States Senate' \\ \midrule
    A1: `who are the two state [senators] of georgia?' \\
    A2: `who are the two [state senators] of georgia?' \\
    A3: Entity not in question \\
\toprule
Question: `who was michael jackson in the wiz?' \\
Entity: `The Wiz (film)' \\ \midrule
    A1: `who was michael jackson in [the wiz]?' \\
    A2: `who was michael jackson in the [wiz]?' \\
\bottomrule
\end{tabular}
\end{center}

\section{Span-Extraction Model for Mention Boundary Detection}
\label{app:span_extraction}
\label{sec:span_extraction}

\newcommand{\bertbase}{\textsc{BERT}$_\mathrm{base}$}

The BLINK Entity Linker requires mention boundaries to be marked in the input. In order to evaluate against BLINK for end-to-end Entity Linking, we train a span-extraction model to first obtain candidate mention boundaries, and subsequently use BLINK on these candidate mentions. Our span extraction model first represents every token $q_i$ in question $q=q_1, \Compactcdots ,q_n$ of length $n$ using a dense representation using \bertbase:
\begin{align}
  \mathbf{q_1}, \Compactcdots ,\mathbf{q_n} = \mathrm{BERT}(q_1, \Compactcdots ,q_n)
\end{align}

The model then computes a start span probability $p_{s}(q_i|q)$ and an end span probability $p_{e}(q_i|q)$ for every token $q_i$ using learnable vectors $\mathbf{w_s}$ and $\mathbf{w_t}$ respectively:
\begin{align}
  p_{s}(q_i|q) = \frac{\exp(\mathbf{w_s}^{\intercal} \mathbf{q_i})}{\sum_j \exp(\mathbf{w_s}^{\intercal} \mathbf{q_j})} \\
  p_{e}(q_i|q) = \frac{\exp(\mathbf{w_e}^{\intercal} \mathbf{q_i})}{\sum_j \exp(\mathbf{w_e}^{\intercal} \mathbf{q_j})}
\end{align}

The model is trained to maximize the likelihood of $p_{s}(q_s|q) \times p_{e}(q_e|q)$ for each correct mention $[q_s, q_e]$ in the training set of WebQSP, and similarly during inference, outputs the top-K scoring spans. These spans are used as mention boundary candidates, to evaluate BLINK in an end-to-end setting.

\section{Analysis}
\label{app:analysis}

\begin{table}[t]
    \centering
    \small
    \begin{tabular}{ccc}
    \toprule
        \textbf{Dataset} & \textbf{\#Examples} & \textbf{\#Agreement} \\
    \midrule
        WebQSP train & 9 & 8 \\
        WebQSP dev & 6 & 5 \\
        WebQSP test & 10 & 10 \\
        GraphQs train & 8 & 8 \\
        GraphQs test & 8 & 8 \\
    \bottomrule
    \end{tabular}
    \caption{Statistics on the shared dataset annotated by all 4 annotators. We count exact-match, unanimous agreements, i.e., both mention boundaries must exactly match, and all 4 annotators must agree.}
    \label{tab:data_agreements}
\end{table}

\paragraph{Training Ablations}
\begin{table}[t]
    \centering
    \small
    \begin{tabular}{rc}
    \toprule
        \textbf{Training method} & \textbf{F1} \\
    \midrule
        Adversarial + Pre-trained candidate encoder & 87.7 \\
        Pre-trained candidate encoder & 46.1 \\
        None (entirely from scratch) & 17.2 \\
    \bottomrule
    \end{tabular}
    \caption{Ablations on our training scheme after 20 epochs.
    Both transfer learning from the pre-trained candidate encoder \textit{and} adversarially training on hard negatives expedite convergence.
    }
    \label{tab:training_ablate}
\end{table}
Table~\ref{tab:training_ablate} presents the contributions of each component of our training scheme to our final result. 
We record model performance on \ourwebq~(valid) after 20 (of 100) epochs of training on Wikipedia, having seen just 20\% of the data. 
We note that both our transfer learning technique and our adversarial hard-negatives training expedites convergence.

\paragraph{Qualitative Error Analysis} We record specific examples for each of our four error categories (technically correct, not enough entities, wrong entities, and insufficient context), detailed in Section~\ref{sec:analysis}.
Note there were 61 total mistakes for~\ourwebq~dev and 158 total mistakes for~\ourgraphq~dev. Specific examples can be found in Table~\ref{tab:qual_examples}.

\section{Implementation Details and Hyperparameters}
\label{app:impl-details}
Following \citet{wu2019zeroshot}, we use BERT$_\text{Large}$ ($\sim$340M parameters) for the question and entity encoder.
The span extraction model detailed in Appendix~\ref{app:span_extraction}, used for our BLINK baselines, is a BERT$_\text{Base}$ model ($\sim$110M parameters).

We lowercase all inputs to ELQ, during both training and inference time, to make it case-insensitive. For both training and inference, the mention scorer considers all spans up to length $L=10$.

\paragraph{Training} During training, we use FAISS~\cite{Johnson_2019} for fast inner product search when mining hard negatives.
We do this in real time, \textit{inside} the training loop.
As we do not update the entity encoder, we were able to train the model with a single FAISS index, greatly increasing the training speed. For further speedup, we use a hierarchical index (\verb|IndexHNSWFlat|), with $efConstruction = 200$ and $efSearch = 256$.

For $\mathcal{L}_\text{ED}$ at each iteration, computing $s(e, [i,j])$ for every $e \in \mathcal{E}$ is intractable. We thus approximate $\mathcal{L}_\text{ED}$ by replacing $\mathcal{E}$ with $\mathcal{E}'$ for the softmax, where $\mathcal{E}'$ is a set of hard negative entities, specifically, negative entities that have the highest 10 similarity scores with the mention representation.

For our~\ourwebq-trained model, we train for up to 100 epochs on~\ourwebq~data, using batch size 128 and context window size of 20 tokens. 
For our Wikipedia-trained model, we split the data evenly into 100 chunks and train on each (thus, making one pass through Wikipedia overall). For Wikipedia, we use batch size 32 and a context window size of 128 tokens.
For Wikipedia +~\ourwebq~model, we take our Wikipedia-trained model and further fine-tune it on~\ourwebq~for up to 100 epochs (using the~\ourwebq~training settings).
For all three training settings, we use the AdamW optimizer with learning rate 1e-5, coupled with a linear schedule with $10\%$ warmup. We clip gradients to max norm $1.0$.

\paragraph{Inference} During inference, we consider all mention candidates $[i,j]$ with mention score $\log p([i,j])\ge \gamma$. If no mention candidate has mention score $\ge \gamma$, we simply take the top-$50$-scoring mentions. $\gamma$ is a threshold we tune on each dataset's dev data.

The linker then retrieves the $10$ closest entity candidates per mention boundary. We use the same hierarchical FAISS index as during training to expedite retrieval. Since the search is approximate, we expect some performance degradation. However, in practice, we found minimal performance degradation for significant speedup. On~\ourwebq~dev set, F1 score decreased from $92.5\to 91.9$, but run-time decreased from $127.0s\to 24.3s$ (for the entire dataset, with batch size 64).

As computing the softmax over all entities for $\log p(e|[i,j])$ is intractable, we simply softmax over our 10 retrieved candidates.
At the end, we threshold the final joint score $\log p([i,j]) + \log p(e|[i,j])$ based on $\gamma$.

\begin{table}[t]
    \centering
    \small
    \begin{tabular}{rcc}
    \toprule
        Model & \ourwebq & \ourgraphq  \\
    \midrule
        ELQ, Wikipedia & $-2.9$ & $-3.5$ \\
        ELQ, Wikipedia +~\ourwebq & $-1.5$ & $-0.9$ \\
    \bottomrule
    \end{tabular}
    \caption{Best threshold hyperparameter value $\gamma$, based on the respective development sets.}
    \label{tab:threshold_hyperparam}
\end{table}

We use manual tuning and binary search to find the best-performing hyper-parameters for the threshold $\gamma$. We optimize for F1-score on the development sets of~\ourwebq and~\ourgraphq. Best settings are reported in Table~\ref{tab:threshold_hyperparam}.

As mention overlaps are not allowed in the questions data, we have an additional global step of removing overlapping mention boundaries --- in the case of multiple entities, we greedily choose the highest-scoring entity each time, and remove all entities which overlap with it.

\section{Infrastructure Details}
We ran all training distributed across 8 NVIDIA TESLA V100 GPUs, each with 32 GB of memory.
For 80-CPU inference, we run on 2 chips of Intel(R) Xeon(R) CPU E5-2698 v4 @ 2.20GHz with 20 cores (40 threads) each.
For 1-CPU inference (reported in Table~\ref{tab:main_result}), we run only on a single core. 

\begin{table*}
    \small
    \centering
    \begin{tabularx}{\textwidth}{rl}
    \toprule
        \textbf{Dataset} & \textbf{Example Error} \\
    \midrule
        & \textbf{Technically Correct (\ourwebq~49.2\%; ~\ourgraphq~23.3\%)} \\
        \ourwebq & what type of \red{guitar} does \blue{john mayer} play? \\
        & \verb|GOLD:| \blue{john mayer}$\to$ ``\verb|John Mayer|" \\
        & \verb|PRED:| \red{guitar}$\to$ ``\verb|Guitar|"; \blue{john mayer}$\to$ ``\verb|John Mayer|" \\
        \\
        \ourwebq & what countries make up \red{continental europe}? \\
        & \verb|GOLD:| \red{continental europe}$\to$ ``\verb|Europe|" \\
        & \verb|PRED:| \red{continental europe}$\to$ ``\verb|Continental Europe|" \\
        \\
    \midrule
        & \textbf{Not Enough Entities (\ourwebq~13.1\%; ~\ourgraphq~51.8\%)} \\
        \ourwebq & what country is the \blue{grand bahama island} in? \\
        & \verb|GOLD:| \blue{grand bahama island}$\to$ ``\verb|Grand Bahama|" \\
        & \verb|PRED:| \\
        \\
        \ourwebq & what \red{children's books} did \blue{suzanne collins} wrote? \\
        & \verb|GOLD:| \red{children's books}$\to$ ``\verb|Children's literature|"; \blue{suzanne collins}$\to$ ``\verb|Suzanne Collins|" \\
        & \verb|PRED:| \blue{suzanne collins}$\to$ ``\verb|Suzanne Collins|" \\
        \\
        \ourgraphq & how many people found \red{o} together? \\
        & \verb|GOLD:| \red{o}$\to$ ``\verb|Oxygen|" \\
        & \verb|PRED:| \\
        \\
        \ourgraphq & the rockets \blue{ares i} and \red{saturn 5} are made by who? \\
        & \verb|GOLD:| \blue{ares i}$\to$ ``\verb|Ares I|"; \red{saturn 5}$\to$ ``\verb|Saturn V|" \\
        & \verb|PRED:| \blue{ares i}$\to$ ``\verb|Ares I|" \\
        \\
        \ourgraphq & where does \red{spirit and opportunity} aim to land? \\
        & \verb|GOLD:| \red{spirit and opportunity}$\to$ ``\verb|Mars Exploration Rover|" \\
        & \verb|PRED:| \\
        \\
    \midrule
        & \textbf{Wrong Entities (\ourwebq~26.2\%; ~\ourgraphq~20\%)} \\
        \ourwebq & which \blue{kennedy} died first? \\
        & \verb|GOLD:| \blue{kennedy}$\to$ ``\verb|Kennedy family|" \\
        & \verb|PRED:| \blue{kennedy}$\to$ ``\verb|John F. Kennedy|" \\
        \\
        \ourwebq & what team did \red{shaq} play for first? \\
        & \verb|GOLD:| \red{shaq}$\to$ ``\verb|Shaquille O'Neal|" \\
        & \verb|PRED:| \red{shaq}$\to$ ``\verb|Tupac Shakur|" \\
        \\
        \ourgraphq & \blue{myuutsu} is what kind of pokemon? \\
        & \verb|GOLD:| \blue{myuutsu}$\to$ ``\verb|Mewtwo|" \\
        & \verb|PRED:| \blue{myuutsu}$\to$ ``\verb|Kyūjutsu|" \\
        \\
        \ourgraphq & in the \red{bart} what kind of trains are used? \\
        & \verb|GOLD:| \red{bart}$\to$ ``\verb|Bay Area Rapid Transit|" \\
        & \verb|PRED:| \red{bart}$\to$ ``\verb|Bart Simpson|" \\
        \\
    \midrule
        & \textbf{Insufficient Context (\ourwebq~11.5\%; ~\ourgraphq~5\%)} \\
        \ourwebq & what was \blue{walt disney}'s first cartoon called? \\
        & \verb|GOLD:| \blue{walt disney}$\to$ ``\verb|The Walt Disney Company|" \\
        & \verb|PRED:| \blue{walt disney}$\to$ ``\verb|Walt Disney|" \\
        \\
        \ourgraphq & what botanical gardens to visit in \red{washington}? \\
        & \verb|GOLD:| \red{washington}$\to$ ``\verb|Washington, D.C.|" \\
        & \verb|PRED:| \red{washington}$\to$ ``\verb|Washington (state)|" \\
        \\
    \bottomrule
    \end{tabularx}
    \caption{Examples of each error type made by our model.}
    \label{tab:qual_examples}
\end{table*}

\end{document}